\documentclass{agujournal2019}
\usepackage{url} 
\usepackage{lineno}
\usepackage{comment}
\usepackage[inline]{trackchanges} 
\usepackage{soul}
\usepackage[english]{babel}
\usepackage{ragged2e}
\usepackage{blindtext}

\linenumbers

\journalname{arXiv cs.AI}
\begin{document}
\nolinenumbers
\justifying
\title{Leveraging Citizen Science for Flood Extent Detection using Machine Learning Benchmark Dataset}

\authors{ Muthukumaran Ramasubramanian\affil{1}, Iksha Gurung\affil{1}, Shubhankar Gahlot\affil{1} \break Ronny Hänsch\affil{2}, Andrew L. Molthan\affil{3}, Manil Maskey\affil{3}}

\affiliation{1}{The University of Alabama Huntsville}
\affiliation{2}{German Aerospace Center}
\affiliation{3}{National Aeronautics and Space Administration}

\correspondingauthor{Muthukumaran Ramasubramanian}{mr0051@uah.edu}

\begin{abstract}

Accurate detection of inundated water extents during flooding events is crucial in emergency response decisions and aids in recovery efforts. Satellite Remote Sensing data provides a global framework for detecting flooding extents.  Specifically, Sentinel-1 C-Band Synthetic Aperture Radar (SAR) imagery has proven to be useful in detecting water bodies due to low backscatter of water features in both co-polarized and cross-polarized SAR imagery. However, increased backscatter can be observed in certain flooded regions such as presence of infrastructure and trees - rendering simple methods such as pixel intensity thresholding and time-series differencing inadequate.  Machine Learning techniques has been leveraged to precisely capture flood extents in flooded areas with bumps in backscatter but needs high amounts of labelled data to work desirably. Hence, we created a labeled known water body extent and flooded area extents during known flooding events covering about 36,000 sq. kilometers of regions within mainland U.S and Bangladesh. Further, We also leveraged citizen science by open-sourcing the dataset and hosting an open competition based on the dataset to rapidly prototype flood extent detection using community generated models. In this paper we present the information about the dataset, the data processing pipeline, a baseline model and the details about the competition, along with discussion on winning approaches. We believe the dataset adds to already existing datasets based on Sentinel-1C SAR data and leads to more robust modeling of flood extents. We also hope the results from the competition pushes the research in flood extent detection further.

\end{abstract}

\section{Plain Language Summary}
We present a machine learning (ML) training dataset containing surface water extents labeled during the time of flood, along with corresponding Sentinel-1 C-band Synthetic Aperture Radar (SAR) imagery. We provide information about the dataset, its detailed data structure, and the data processing procedures involved in generating the dataset. We also incorporated citizen science by hosting an open-for-all competition based on the dataset. We hope this dataset augments other publicly available flood extent datasets and help in creating better flood extent detection models. We also share the details on the competition winning models to foster innovative ideas on improving flood extent detection.

\section{Introduction}
Floods are major natural disasters and contribute to widespread property damage, loss of agricultural productivity, loss of lives, displacement of those affected, and long-term socioeconomic consequences \cite{dawson2009integrated, boros2014long, long2014flood,inambao2013namibia}. Knowing the spatial extent of floods is crucial for federal agencies, local authorities, and nonprofits in providing emergency procedures and disaster relief. 
Remote sensing has been used extensively in the community to monitor these events \cite{sanyal2004application, schumann2009progress, jain2005delineation, klemas2015remote}. The temporal and spatial availability of remote sensing data provided by recent governmental and commercial satellites enable the community to make large scale analysis of flood events with greater detail than ever before. Spectral and radio back-scatter remote sensing data has been widely used \cite{anusha2020flood, mateo2021towards,helleis2022sentinel, schmitt2019sen12ms, bonafilia2020sen1floods11}. Water extents are easily identifiable spectral data but clouds pose significant challenge when using the optical data for detecting floods due to occlusion. Probability of clouds present over flooding regions are also high in tropical areas that are flood prone, as well as at around the time of flooding due to high precipitation. Active sensors such as Synthetic Aperture Radar (SAR) can monitor earth surface through clouds and aerosols and are independent of solar illumination. These properties make them well suited for monitoring flooding events through cloud cover and in both day and night. 

The Sentinel-1 satellite consists of two polar orbiting satellites housing C-Band SAR sensor. \cite{TORRES20129}. the constellation has a combined revisit time of 6 days and 10m spatial resolution. Since surface water typically correspond to low back-scatter values, thresholding on a single image, or observing changes in series of images \cite{nhess-9-303-2009, SCHLAFFER201515} is sufficient to detect water extents. This can be used over flooded areas to identify the extent affected by calculating difference between water areas during flood and known water areas. However presence of building and vegetation footprints amongst the flooded regions pose a challenge in identifying flood extents due to higher back-scatter values arising from the "double bounce" effect. These regions exhibit "speckles" or points of higher back-scatter values amongst low back-scatter values that correspond to water body. Due to this effect, oftentimes thresholding is followed by post-processing the resulting flood extent maps. However, inaccuracies still persist, especially in identifying isolated streams of water body\cite{SHEN2019302}.

Deep learning has been widely used in the earth science community for identifying features such as clouds, land cover, and building footprints due to their superior performance compared to traditional algorithms \cite{rs12020207}. Convolutional Neural Networks (CNN) \cite{10.1145/3065386} are especially used for pixel-wise image segmentation due to their state-of-the-art performance. U-Net \cite{ronneberger2015unet} is a CNN based model that has been used with great success in image segmentation tasks that use earth observation data \cite{rs12020207, russwurm2018multi, hwang2018automatic, yi2019semantic}. This architecture uses encoder-decoder architecture and skip-connections between the layers to downsample image into a lower, dense representation and then upsample to predict pixel-wise classes. They perform well with satellite image inputs as they capture the contextual information that exist thoughout the image while retaining the high-frequency, localized perceptual information. Recent works in flood extent detection use U-Nets with good success \cite{mateo-article, rs12162532}. 

However, training any deep learning model require high quality labeled datasets, with an emphasis on datasets with high spatio-temporal variety. Datasets specifically aimed at developing flood extent segmentation exist \cite{bonafilia2020sen1floods11, schmitt2019sen12ms} and we aim to add to the data, with focus on flooding in the U.S inland waters (Alabama and Missouri Region). Our motivation to create such a dataset is two-fold: 1. Aid in creating high-quality models for monitoring flooding within U.S inland regions covered in the dataset. 2. Augment existing datasets that lead to better generalization of global-scale flood segmentation models.

Motivated by the success of U-Nets, we used the architecture to create a baseline model for detecting flood extents. Details on the model and its performance will be discussed in section \ref{baseline}. However, finding optimal machine learning solutions is an exhaustive process in itself, mostly involving trial-and-error experimenting of various model architectures. Citizen science has been used extensively to find the best solution for problems in both scientific and commercial sectors \cite{beaumont2014milky, borne2011zooniverse}. As part of incorporating citizen science and involving the broader science community to help find innovative and accurate solutions, we open-sourced the processed flood extent dataset in a format that can be used by data scientists from all disciplines and sectors. Furthermore, we designed and hosted a competition on flood extent detection using the aforementioned dataset to accelerate use of innovative ML techniques for flood extent detection. The competition is showcased by the International Conference on Emerging Techniques in Computational Intelligence (ICETCI), 2021.

The details of the flood extent dataset are explained in the following section. 

\section{Overview of the Dataset}

\subsection{Data Collection}
SAR imagery for various flood events were acquired from the European Space Agency (ESA) Sentinel-1A and Sentinel-1B missions, offering C-band, dual polarized (co-pol VV and cross-pol VH) imagery for a number of known flood events of interest for regions within the United States (Alabama, Nebraska and Iowa), and one within Asia (Bangladesh), as shown in the Table \ref{tab:sources}. The images are acquired using Interferometric Wide Swath mode \cite{1677745}, which is the main acquisition mode over land. 

\begin{table}[]
\scriptsize
\begin{tabular}{lllll}
\textbf{Location} & \textbf{Date-time(UTC)} & \textbf{Bounding Box} & \textbf{Area} & \textbf{Water/Non-water Pixels} \\North Alabama, USA & 2019-06-30 23:46:55 & (-88.31, 33.73), (-85.19, 35.67) & 606.58 & 884666/63084481 \\North Alabama, USA & 2019-07-24 23:46:57 & (-88.31, 33.73), (-85.19, 35.67) & 606.73 & 780321/63201432 \\North Alabama, USA & 2019-06-18 23:46:54 & (-88.31, 33.73), (-85.19, 35.67) & 606.58 & 899471/63069676 \\North Alabama, USA & 2019-11-21 23:47:00 & (-88.31, 33.73), (-85.19, 35.67) & 606.71 & 695265/63286488 \\North Alabama, USA & 2019-10-04 23:47:00 & (-88.31, 33.73), (-85.19, 35.67) & 606.63 & 876931/63094671 \\North Alabama, USA & 2019-12-27 23:46:59 & (-88.31, 33.73), (-85.19, 35.67) & 606.72 & 891660/63090093 \\North Alabama, USA & 2019-09-10 23:46:59 & (-88.31, 33.73), (-85.19, 35.67) & 606.73 & 878451/63103302 \\North Alabama, USA & 2019-06-06 23:46:54 & (-88.31, 33.73), (-85.19, 35.67) & 606.58 & 922979/63046168 \\North Alabama, USA & 2019-08-05 23:46:58 & (-88.31, 33.73), (-85.19, 35.67) & 606.73 & 878599/63103154 \\North Alabama, USA & 2019-05-25 23:46:53 & (-88.31, 33.73), (-85.19, 35.67) & 606.72 & 930753/63051000 \\North Alabama, USA & 2019-04-19 23:46:52 & (-88.31, 33.73), (-85.19, 35.67) & 606.58 & 914037/63055110 \\North Alabama, USA & 2019-05-13 23:46:53 & (-88.31, 33.73), (-85.19, 35.67) & 606.72 & 934468/63047285 \\ North Alabama, USA & 2019-08-29 23:46:59 & (-88.31, 33.73), (-85.19, 35.67) & 606.58 & 907680/63061467 \\North Alabama, USA & 2019-03-02 23:46:51 & (-88.31, 33.73), (-85.2, 35.67) & 603.49 & 954406/62717042 \\Red River North, Alabama, USA & 2019-01-28 00:22:46 & (-100.03, 47.04), (-96.08, 49.01) & 779.37 & 327272/60057282 \\Red River North, Alabama, USA & 2019-06-21 00:22:50 & (-100.03, 47.04), (-96.08, 49.01) & 779.31 & 616778/59767777 \\Red River North, Alabama, USA & 2019-05-16 00:22:48 & (-100.03, 47.04), (-96.08, 49.01) & 779.31 & 561353/59823202 \\Red River North, Alabama, USA & 2019-04-22 00:22:47 & (-100.03, 47.04), (-96.08, 49.01) & 779.32 & 382859/60001696 \\Red River North, Alabama, USA & 2019-06-09 00:22:49 & (-100.03, 47.04), (-96.08, 49.01) & 779.32 & 619382/59765173 \\Red River North, Alabama, USA & 2019-05-04 00:22:47 & (-100.03, 47.04), (-96.08, 49.01) & 779.32 & 589922/59794633 \\Red River North, Alabama, USA & 2019-02-09 00:22:46 & (-100.03, 47.04), (-96.08, 49.01) & 779.18 & 304670/60068896 \\Red River North, Alabama, USA & 2019-05-28 00:22:48 & (-100.03, 47.04), (-96.08, 49.01) & 779.32 & 826708/59557847 \\Red River North, Alabama, USA & 2019-04-10 00:22:46 & (-100.03, 47.04), (-96.08, 49.01) & 779.21 & 479849/59893717 \\Red River North, Alabama, USA & 2019-01-04 00:22:47 & (-100.03, 47.04), (-96.08, 49.01) & 779.37 & 379111/60005443 \\Red River North, Alabama, USA & 2019-03-17 00:22:46 & (-100.03, 47.04), (-96.08, 49.01) & 779.32 & 463005/59921550 \\Red River North, Alabama, USA & 2019-01-16 00:22:47 & (-100.03, 47.04), (-96.08, 49.01) & 779.49 & 525194/59870351 \\Red River North, Alabama, USA & 2019-02-21 00:22:46 & (-100.03, 47.04), (-96.08, 49.01) & 779.21 & 241432/60132134 \\Florence, Alabama, USA & 2018-05-22 23:13:44 & (-80.0, 33.17), (-76.8, 35.21) & 653.1 & 794112/69374988 \\Florence, Alabama, USA & 2018-06-15 23:13:45 & (-80.0, 33.17), (-76.8, 35.21) & 653.09 & 579446/69589654 \\Florence, Alabama, USA & 2018-08-02 23:13:48 & (-79.99, 33.17), (-76.8, 35.21) & 653.09 & 619835/69549265 \\Florence, Alabama, USA & 2018-09-19 23:13:50 & (-80.0, 33.17), (-76.8, 35.21) & 653.18 & 926438/69253135 \\Florence, Alabama, USA & 2018-10-01 23:13:50 & (-80.0, 33.17), (-76.8, 35.21) & 653.1 & 855325/69313775 \\Florence, Alabama, USA & 2018-06-03 23:13:44 & (-79.99, 33.17), (-76.8, 35.21) & 653.09 & 686080/69483020 \\Florence, Alabama, USA & 2018-05-10 23:13:43 & (-80.0, 33.17), (-76.8, 35.21) & 653.1 & 754570/69414530 \\Florence, Alabama, USA & 2018-09-07 23:13:50 & (-80.0, 33.17), (-76.8, 35.21) & 653.09 & 750525/69418575 \\Florence, Alabama, USA & 2018-07-21 23:13:47 & (-80.0, 33.17), (-76.8, 35.21) & 653.1 & 859347/69309753 \\Florence, Alabama, USA & 2018-07-09 23:13:46 & (-79.99, 33.17), (-76.8, 35.21) & 653.09 & 719576/69449524 \\Missouri River Basin , USA & 2017-12-10 00:21:19 & (-98.02, 39.56), (-94.54, 41.61) & 716.49 & 132365/68326168 \\Missouri River Basin , USA & 2017-11-16 00:21:20 & (-98.02, 39.56), (-94.54, 41.61) & 716.31 & 177903/68259092 \\Missouri River Basin , USA & 2017-06-01 00:21:14 & (-98.07, 39.56), (-94.53, 41.78) & 785.8 & 206517/75982683 \\Missouri River Basin , USA & 2017-05-20 00:21:13 & (-98.02, 39.56), (-94.54, 41.61) & 716.41 & 154545/68293219 \\Missouri River Basin , USA & 2017-09-29 00:21:20 & (-98.02, 39.56), (-94.54, 41.61) & 716.41 & 178555/68269209 \\Missouri River Basin , USA & 2017-03-09 00:21:10 & (-98.02, 39.56), (-94.54, 41.61) & 716.3 & 2763/68434232 \\Missouri River Basin , USA & 2017-04-02 00:21:11 & (-98.02, 39.56), (-94.54, 41.61) & 716.31 & 188260/68248735 \\Missouri River Basin , USA & 2017-12-22 00:21:18 & (-98.02, 39.56), (-94.54, 41.61) & 716.4 & 160240/68287524 \\Missouri River Basin , USA & 2017-05-08 00:21:13 & (-98.02, 39.56), (-94.54, 41.61) & 716.31 & 168175/68268820 \\Missouri River Basin , USA & 2017-01-08 00:21:12 & (-98.01, 39.59), (-94.55, 41.64) & 710.36 & 17685/67888491 \\Missouri River Basin , USA & 2017-02-13 00:21:21 & (-98.19, 40.2), (-94.67, 42.25) & 720.88 & 217719/67615435 \\Bangladesh & 2017-03-14 11:56:09 & (90.54, 23.54), (93.74, 27.01) & 1109.15 & 2326553/136686815 \\Bangladesh & 2017-07-12 11:56:15 & (90.54, 23.54), (93.74, 27.01) & 1109.37 & 3069449/135968445 \\Bangladesh & 2017-06-06 11:56:13& (90.54, 23.54), (93.74, 27.01) & 1109.05 & 2521135/136479970 \\
\end{tabular}
\caption{Summary of labeled flood events and the respective regions}
\label{tab:sources}
\end{table}


\subsection{Data Processing}
\begin{figure}[]
        \centering
        \includegraphics[width=0.5\textwidth]{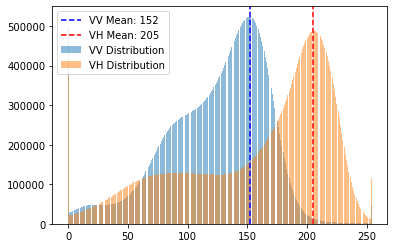}
        \caption{Averaged Histogram of 52 processed VV and VH GeoTiffs}
        \label{fig:dist}
    \end{figure}

    Following the data collection, SAR VV, VH images were processed to a radiometric and terrain-corrected (RTC) image of the radar amplitude, then converted to a grayscale image for visual analysis using the Hybrid Pluggable Processing Pipeline or ”HyP3” system \cite{HyP3}. HyP3 takes the Sentinel archive and creates a set of processes to get to a consistent method of generating the VV / VH amplitude or power imagery. 
    The imagery is then preprocessed by normalizing and enhancing the contrast for better visual identification of features. Finally, the images are converted to 0-255 grayscale images. A total of 52 labeled GeoTiff files are converted into pairs of grayscale images: one for VV and one for VH polarization. The average data distribution statistics for VV and VH geotiff images is shown in Fig. \ref{fig:dist}

\subsection{Data Labeling}
    Once the "HyP3" pipeline converts the SAR imagery corresponding to the flood events that were collected (Table \ref{tab:sources}) into single-band GeoTiffs, one each for VV and VH polarization, The GeoTiffs are made available in a collaborative labeling tool called ImageLabeler \cite{ImageLabeler}. This tool provides a map interface with interchangeable layers for visual identification of floods. It also facilitates simultaneous, colloborative labeling of flooded extents that made the laborious task of analysing 36,000 sq.km of data fast and easy. Detailed polygons engulfing the water features were drawn for suspected water areas and vetted through multiple visual inspections with SAR imagery experts. Here, emphasis was on the labeling of two characteristic features of SAR data: 1. Open water areas where specular reflection of the radar signal off of the relatively still, flat open water surface results in reduced backscatter, low amplitude, and an overall darkened appearance within the image. In normal conditions, ponds, lakes, and rivers will appear dark and usually include crisp edges where water adjoins the nearby vegetation and topography. 2. Additional dark - and sometimes non-contiguous features due to double-bounce effect that occur following heavy rains and flooding and often include expanded, flooding growth of dark regions along the normal water areas or standing water in fields or other topographic features where ponding of water is likely \cite{liang2020local, doi:10.1080/01431160116902}. Emphasis was made on the labeling of these two features for generating the training dataset. We believe inclusion of the latter feature with the dataset poses harder challenge compared to open water features in flood extent modeling and pushes researchers to leverage more sophisticated ML techniques.
     
    Areas that were “dark” in the SAR images and might not have been water bodies were particularly challenging to examine. In such cases, Spectral data such as Moderate Resolution Spectrometer (MODIS) were used to make sure that they were permanent water bodies and not inundated. These polygons represent both the open water and flood extent class as expert labels and were used to classify open water pixels relative to vegetation and other classes in the image. 


\begin{figure}[ht]
        \centering
        \includegraphics[width=0.9\textwidth]{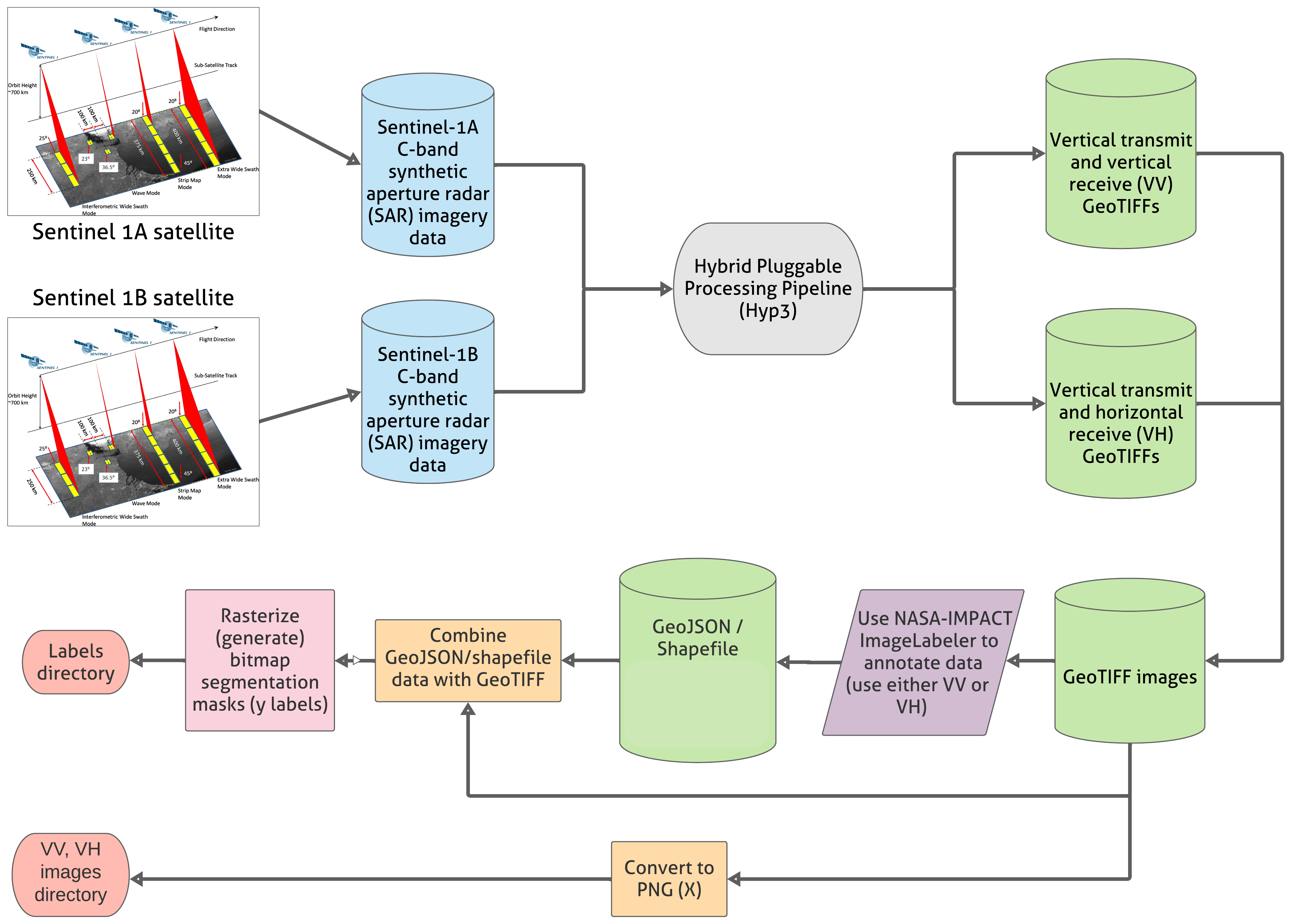}
        \caption{Data pre-processing \& generation workflow}
        \label{fig:data_generation}
    \end{figure}


\subsection{ML Training Data Preparation}
The VV, VH grayscale images are split into into $256\times256$ tiles (scenes) with a per-pixel resolution of 30 meters approximately. The tiles with nodata values and tiles outside of the valid SAR boundary are eliminated. Tiles without any presence of a water body is also discarded. In addition to the flood extents, World Water Bodies GeoTiff data from UCLA Geoportal \cite{wwb} is also processed into water body rasters and provided for the respective regions(see \ref{fig:data_generation}). This inclusion of permanent water body in the training data is to aid the model learn inundation-specific features. 

\subsection{Data Stratification} 
\label{data:split}
The data is divided into train, validation, and test set. They are geographically divided to make sure that the train-validation and test distributions does not have data leakage issues. \textit{Nebraska, North Alabama, Bangladesh and Florence, AL} regions are used for the train and validation sets whereas the \textit{Red River North} region is used for the test set.

\begin{figure}
    \centering
    \includegraphics[width=0.9\textwidth]{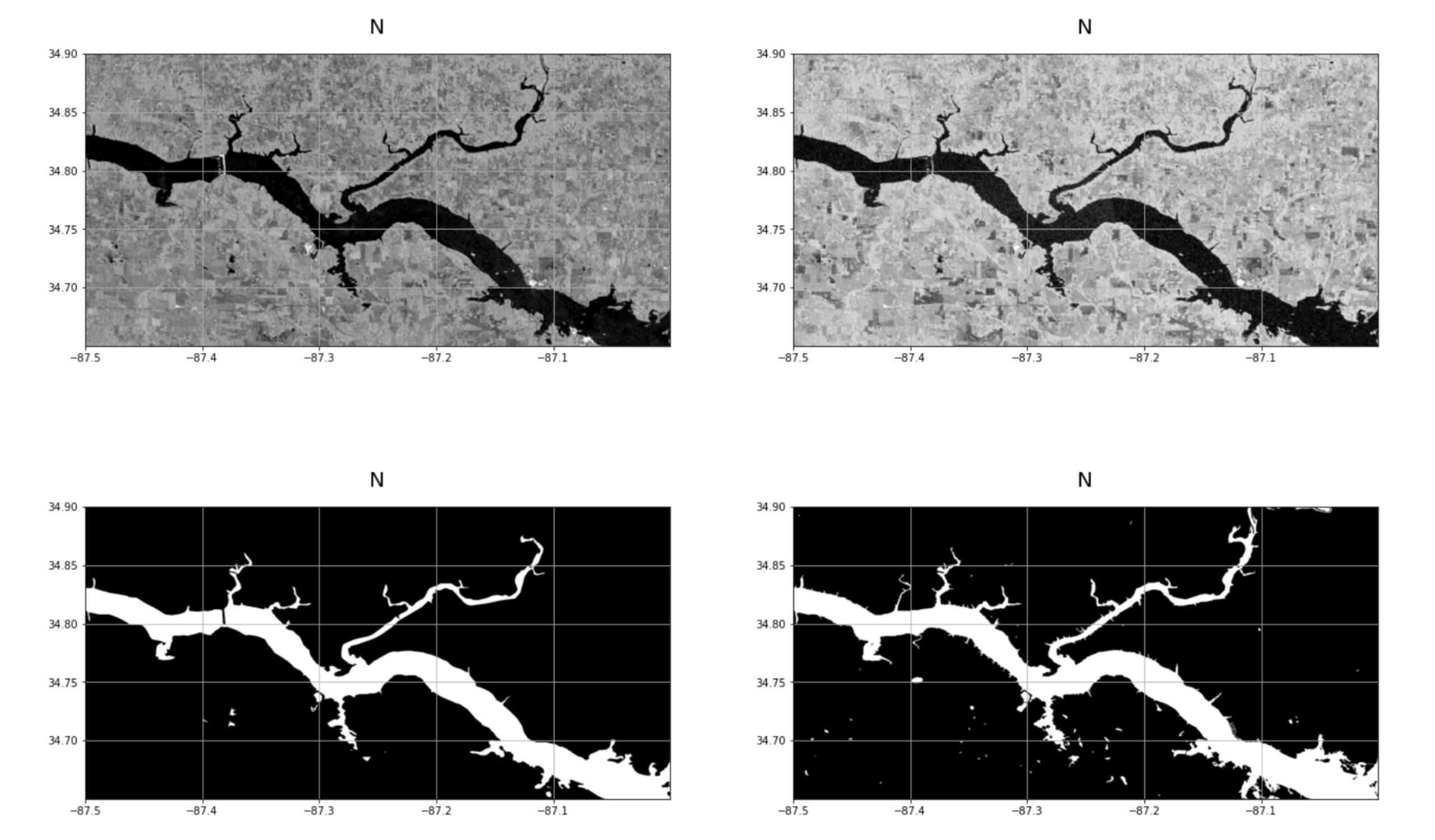}
    \caption{HyP3 Processed SAR Geotiff Imagery and Labels during flood event in northern Alabama, dated 06-30-2019. From left to right and top to bottom : VV, VH, Permanent Water Body label, Flood label}
    \label{fig:SAR images}
\end{figure}

Each training dataset contains four images as described below:
\begin{itemize}
    \item VV (polarization amplitude) (Fig.~\ref{fig:SAR images} top left)
    \item VH (polarization amplitude) (Fig.~\ref{fig:SAR images} top right)
    \item Permanent water body label (Fig.~\ref{fig:SAR images} bottom right)
    \item Flood extent label (Fig.~\ref{fig:SAR images} bottom left)
\end{itemize}

Spatio-temporal information in the data is removed and dataset is anonymised to support hosting the Emerging Techniques in Computational Intelligence (ETCI) Flood extent detection competition, which will be discussed in next section. This dataset was made available in MLhub data repository \cite{dataset_DOI} along with spatio-temporal information after the competition was concluded.

\section{Flood Extent Modeling and Citizen Science}

In this section, we discuss a simple baseline model trained on the anonymised dataset. A metric to benchmark the performance of the models is also defined. Further, we also provide details on incorporating citizen science by the way of open-for-all competition and conclude with a discussion on the top three ranked models (based on the metric). 



\subsection{Baseline Metric}
\label{baseline}
We establish the baseline metric as the Intersection Over Union (IOU) scoring function, also known as Jaccard Index given by Eq.~\ref{eq:1} for all the models discussed in the following sections. 

\begin{equation}\label{eq:1}
    J(A, B) = \frac{|A \cap B|}{|A \cup B|}
\end{equation}
where $A$ and $B$ are the model estimated and reference flood masks respectively. In simple terms, IOU is the area of overlap between the predicted segments and ground truth and the union between them. The metric ranges from 0-1, where 0 signifies no overlap, and 1 signifies full overlap.

\subsection{Baseline model}
A U-Net model is trained on the flood extent dataset using the training split as described in section \ref{data:split}. The model used a \textit{ResNet50}\cite{he2015deep} backbone and the network was trained for 100 epochs without any regularization by stacking VV, VH and Water body label images as 3-channels of the input tensor. 

\begin{figure}
        \centering
        \begin{minipage}{0.5\textwidth}
            \centering
            \includegraphics[width=0.55\textwidth]{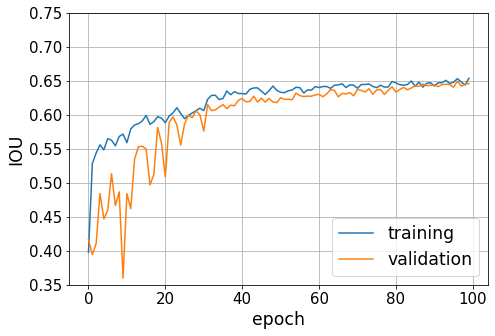}
            \caption{Train-validation scores for UNet}
            \label{fig:unet train-val}
        \end{minipage}\hfill
        \begin{minipage}{0.5\textwidth}
            \centering
            \includegraphics[width=1.1\textwidth]{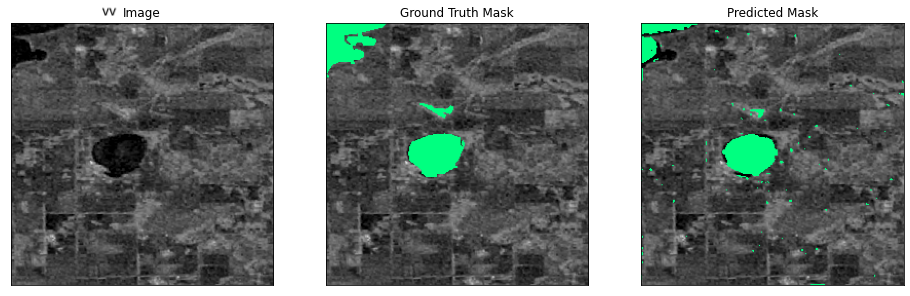}
            \caption{Sample U-Net prediction}
            \label{fig:unet_pred}
        \end{minipage}\hfill
    \end{figure}

Fig. \ref{fig:unet train-val} shows the IOU scores on training and validation splits versus the epochs. As shown in the graph, The training is initially noisy, with significant gaps between validation split and training split IOU scores. We hypothesize this could be due to geographical differences between the splits as the model is struggling to generalize in the early stages. By the end of 100 epochs, the two scores stabilize to similar IOU scores, indicating good generalization. The IOU scores on the test set (as described in Section \ref{data:split}) is \textbf{0.6198}. Fig. \ref{fig:unet_pred} shows a sample from U-Net predictions. The image on the left is the VV image and the image in the center is the ground truth flood mask, and on the right is the predicted flood mask overlaid on top of the actual image. The baseline models provide a starting point for more advanced deep learning techniques. 
    
\subsection{Citizen Science for Flood extent detection}

 With our goal to incorporate citizen science for flood extent detection in mind, we set to get the larger ML community involved by designing an accessible, open-for-all competition; The flood extent detection competition was developed in collaboration with the International Conference on Emerging Techniques in Computational Intelligence (ICTE) and Geoscience and Remote Sensing Society (GRSS). The competition commenced on April 15, 2021 and concluded on July 15, 2021. The competition received a total of 194 submissions, made by 16 teams. The competition was hosted using the Codalab competition platform \cite{codalab_2021} \cite{etci_gitpages} which is an open source, community driven data science competition platform.\\

The competiton was organized into two phases as illustrated: \\
\textbf{Phase 1 (Development):} Participants were provided with training set (which includes reference labels) and validation set (without reference labels until phase 1 concludes) to train and validate their algorithms. Note that the splits indicated here are the same as illustrated in section \ref{data:split}. Participants can submit prediction results for the validation set to the codalab competition website to get feedback on the performance from April 15 to May 14, 2021. The performance of the best submissions from each account were displayed on the leaderboard. \\
\textbf{Phase 2 (Test):} Participants received the reference labels for the validation set for model tuning and test dataset (without the corresponding reference labels) to generate predictions and submit their binary classification maps in array format from May 15 to June 30, 2021. After evaluation of the results, three winners were announced on July 1, 2021. \\

\begin{table}[]
    \tiny
    \begin{tabular}{|l|l|l|l|}
    \hline
    \textbf{Place} & \textbf{Participants}                                             & \textbf{Model Architecture}          & \textbf{IOU} \\ \hline
    1              & Haoran Xu, Bingheng Li, Yong Zheng, Yujie Liu (Xidian University) & U-Net model with Self-Attention      & 0.7681       \\ \hline
    2              & Sayak Paul, Siddha Ganju                                          & U-Net Ensemble with weak supervision & 0.7654       \\ \hline
    3              & Shagun Garn, Binayak Ghosh, Mahdi Motagh                          & Feature Pyramid Network              & 0.7506       \\ \hline
    \end{tabular}
    \caption{ETCI 2021 Competition on Flood Extent Detection Winners and Winning Model Architectures}
    \label{tab:comp}
\end{table}

Three winners were chosen based on their phase 2 test IOU score, shown in Table \ref{tab:comp}.  The highest test IOU score achieved was 0.7681 followed by 0.7654 and 0.7506. The winning model (Haoran Xu et al, Xidian University) (unpublished) used a U-Net architecture with a transformer based "self-attention" \cite{vaswani2017attention} layer deployed at the bottleneck layer, as shown in Fig. \label{fig:winner}. By doing so, The model was able to effectively capture long-range dependencies that exist among local features in the bottleneck layer.   \\

\begin{figure}[]
        \centering
        \includegraphics[width=0.9\textwidth]{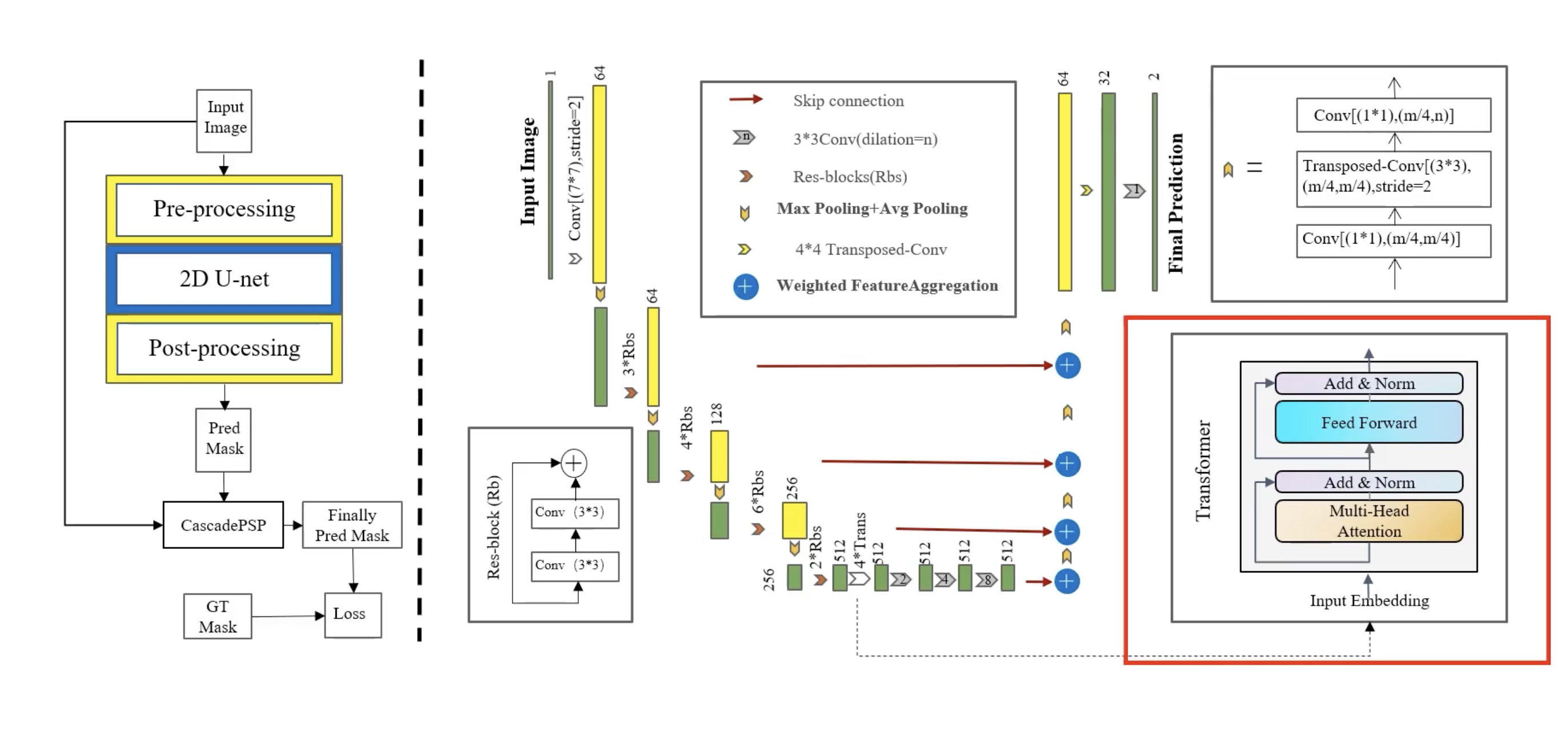}
        \caption{U-Net Model with Self-attention in bottleneck layer (Xu et al, Xidian University)}
        \label{fig:winner}
    \end{figure}
\begin{figure}[ht]
        \centering
        \includegraphics[width=0.9\textwidth]{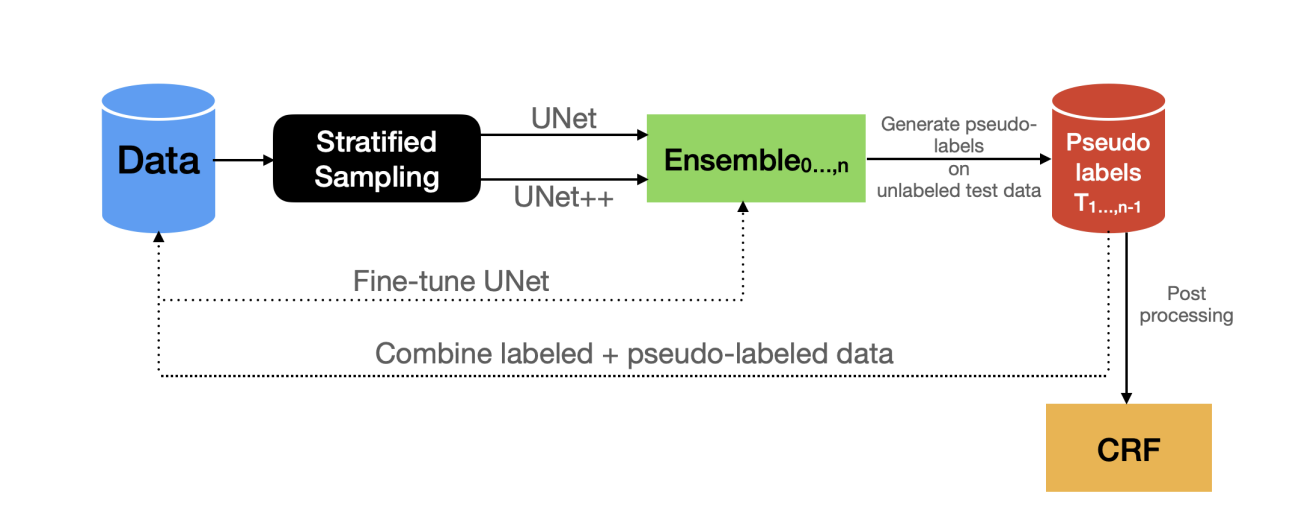}
        \caption{U-Net Model Ensemble and Weak Supervision by generating pseudo-labels \cite{paul2021flood}}
        \label{fig:second}
    \end{figure}

The second place winner\cite{paul2021flood} used an ensemble of U-Net architectures, and leveraged weak supervision \cite{lakshminarayanan2016simple} to generate high quality "pseudo-labels" to augment the high-confidence human annotated labels. The architecture is given in Figure. \ref{fig:second}. \\

The Third place winner\cite{Garg_2021} used a standard Feature Pyramid Network ~(FPN)\cite{lin2016feature} with an efficientnet-B17\cite{tan2019efficientnet} as the encoding model to obtain test IOU score of 0.75. A summary of the winning models and participants are given in Table \ref{tab:comp}. \\

While most of the models submitted to the competition (as well as the baseline model) used well known variations of CNNs such as U-Nets and FPN, The best performing models incorporated additional innovative techniques such as self-attention and weak supervision. These differences amounted to a significant 15 percent increase in model performance compared to baseline, even when the underlying model architecture remained the same. 

\section{Conclusion}
 The purpose of this work is to enable rapid prototyping of high quality machine learning models for the task of flood extent detection. We started by creating labels for flood extents using Sentinel-1 SAR images and made the dataset easily accessible to the broader ML community. By doing so, we were able to design a competition to get the ML community involved in the problem of flood extent detection. Our contributions with this work is two-fold.; First, We created a high quality flood extent dataset that can be used to improve existing capabilities. Second, and more importantly, we believe we were able to collect multiple innovative ideas and solutions that push the limits of flood extent detection within a short span of time by incorporating citizen science. 

 \section{Open Research}
 The dataset for flood extent detection is made available via this unique identifier: \url{10.34911/RDNT.EBK43X}\cite{dataset_DOI} [CC-BY-4.0]. The dataset is also accessible through Radiant MLHub data registry \url{https://registry.mlhub.earth/10.34911/rdnt.ebk43x}. 

\acknowledgments
We would like to thank the NASA-IMPACT, IEEE GRSS, University of Alabama in Huntsville, University of Alaska Fairbanks (UAF), NASA Disaster Team at Marshall Space Flight Center (MSFC) for supporting training data generation and International Conference on Emerging Techniques in Computational Intelligence (ICETCI) for hosting the competition. Special thanks to Dr. Franz J. Meyer at UAF for his help with the data generation, Dr. Ronny H{\"a}nsch for his support throughout the competition and chairing this event at ICETCI, and the students: Jacob Robinson, Kiahna Mollette, Kaitlyn Wheeler, Stefanie Mehlich and Zachary Helton; and research staff Ankur Shah and Ronan M. Lucey for their help with data curation.

\bibliography{agusample}

\end{document}